\pdfoutput=1

\documentclass[11pt]{article}

\usepackage[]{acl}

\usepackage{times}
\usepackage{latexsym}
\usepackage{amsmath} 

\usepackage[T1]{fontenc}

\usepackage[utf8]{inputenc}

\usepackage{microtype}

\usepackage{inconsolata}

\usepackage{graphicx}

\usepackage{multirow}

\usepackage{listings}  

\usepackage{caption}
\usepackage{subcaption}

\title{AILS-NTUA at SemEval-2025 Task 8: Language-to-Code prompting and Error Fixing for Tabular Question Answering}

\author{Andreas Evangelatos, Giorgos Filandrianos, Maria Lymperaiou, \\  \textbf{Athanasios Voulodimos, Giorgos Stamou}
   \\
National Technical University of Athens
\\ \texttt{\href{mailto:andreasevagelatos@mail.ntua.gr}{andreasevagelatos@mail.ntua.gr},  \{\href{mailto:geofila@islab.ntua.gr}{geofila}, \href{mailto:marialymp@islab.ntua.gr}{marialymp}\}@ails.ece.ntua.gr,} \\
\texttt{\href{mailto:thanosv@mail.ntua.gr}{thanosv@mail.ntua.gr},
\href{mailto:gstam@cs.ntua.gr}{gstam@cs.ntua.gr}}\\
}

\begin{document}
\maketitle
\begin{abstract}
In this paper, we present our submission to SemEval-2025 Task 8: Question Answering over Tabular Data. This task, evaluated on the DataBench dataset, assesses Large Language Models' (LLMs) ability to answer natural language questions over structured data while addressing topic diversity and table size limitations in previous benchmarks. We propose a system that employs effective LLM prompting to translate natural language queries into executable code, enabling accurate responses, error correction, and interpretability. Our approach ranks first in both subtasks of the competition in the proprietary model category, significantly outperforming the organizer's baseline.
\end{abstract}

\section{Introduction}
The integration of Large Language Models (LLMs) into question-answering (QA) systems over tabular data has garnered significant attention in recent research, due to their capability to translate natural language into structured queries \cite{fang2024largelanguagemodelsllmstabular}. First attempts in language-to-query conversion rely on fine-tuning pre-trained language models, so that they acquire tabular understanding \cite{liu2022tapextablepretraininglearning}. The vast contextual knowledge of LLMs, as well as their exposure on multiple modalities, including tabular comprehension, enables them to implicitly translate natural language in structured queries. As tabular QA
arises as an emergent LLM ability \cite{chen-2023-large}, more and more prompting-based methods showcase improvements \cite{Hegselmann2022TabLLMFC, Nam2024TabularTL}.

Evaluating LLMs as tabular reasoners was explored with the DataBench dataset \cite{oses-grijalba-etal-2024-question}, comprising 65 datasets from various domains, accompanied by manually crafted questions in natural language. Answers fall into different categories, such as boolean, categorical, numerical or list. Experiments utilizing in-context learning and code-based prompting reveal that while LLMs show promise, there is significant room for improvement, particularly in smaller scales.

In this paper, we explore a wide range of prompting strategies for Large LLMs to identify optimal methods for converting natural language queries into code and effectively interacting with tabular data.
Specifically, we propose a system for generating executable Python functions from natural language queries through LLM prompting. Our approach achieves accuracy scores of $85.63\%$ and $87.93\%$ in the DataBench and DataBench Lite tasks respectively, ranking first in both tasks in the proprietary models category. After human evaluation, the accuracy increases to $89.85\%$ and $88.89\%$ in DataBench and DataBench Lite respectively. In our system, LLMs are explicitly instructed to articulate their reasoning process, improving the reliability of their outputs and facilitating interpretability.

Our code is available on GitHub\footnote{\url{https://github.com/andrewevag/SemEval2025-Task8-TQA}}.

\section{Background}
\subsection{Task description}

The data used for this task are sourced from DataBench \cite{oses-grijalba-etal-2024-question}. The task consists of queries in the form of $(T, Q)$ where $T$ is a structured table containing a set of $c$ columns $\{ \mathrm{col}_{1}, \dots,\mathrm{col}_{c} \}$ and a set of $r$ rows $\{ \mathrm{row}_{1},\dots,\mathrm{row}_{r} \}$. Each row consists of $c$ values $\{\mathrm{row}_{i,1}, \dots, \mathrm{row}_{i,c}\}$, which correspond to the entries in the table for the respective columns. These values can be of various data types commonly found in real-world datasets in English, such as numbers, categorical values, dates, and text. Additionally, some values may be missing. The expected answer to the query is a value $a$ of type 
\texttt{boolean, category, number, list[category]} or, \texttt{list[number]}. 
Answers are computed based on a small subset of the columns and may either be values directly present in $T$ or include statistics computed on these values.

Two subtasks are proposed with regards to the size of the table $T$ of the input query:

 \textbf{Subtask I: DataBench QA} – A table of any size is provided along with a question in natural language. 
 
 \textbf{Subtask II: DataBench Lite QA} – The same task is performed using a sampled version of the tables, where a maximum of 20 rows is retained.

\subsection{Related work}

LLMs excel in various tasks but struggle with tabular data, especially when key information is dispersed across large tables and complex queries \cite{10.1145/3539618.3591708}. This necessitates shifting from direct LLM reasoning to generating intermediate representations in SQL or Python \cite{zhang2023reactableenhancingreacttable}. For instance, \citet{zhang2023bridginggapdecipheringtabular} serialize table schemas, enabling GPT-3 to iteratively refine SQL queries by correcting syntax and compilation errors. Similarly, Plan-of-SQLs (POS) \citep{giang2024interpretablellmbasedtablequestion} decomposes complex queries, reducing reliance on advanced Text-to-SQL models. PAL \cite{gao2023palprogramaidedlanguagemodels} generates intermediate programmatic steps, outsourcing execution to a Python interpreter.
\citet{lei2023tableqakitcomprehensivepracticaltoolkit} integrate multiple LLMs into a toolkit, leveraging Chain of Thought (CoT) \cite{wei2023chainofthoughtpromptingelicitsreasoning} and Program of Thoughts (PoT) \cite{chen2022program} prompting to enhance SQL-based reasoning. TabLLM \cite{zha2023tablegptunifyingtablesnature} further extends comprehension by jointly training LLMs on tables and text. Building on this, \citet{10.1145/3539618.3591708} improve table-based reasoning by segmenting large tables into relevant sub-tables and decomposing complex queries.
Unlike intermediate representation-based methods, \citet{wu-hou-2025-efficient} use in-context learning with retrieval-augmented generation (RAG) for direct tabular query answering. \citet{agarwal2025hybridgraphstableandtextbased} enhance in-context learning by integrating tabular data and queries into a unified hybrid graph. \citet{10.1145/3677052.3698685} employ a step-wise pipeline to distill knowledge, enabling smaller models for discrete reasoning over tabular data.

\begin{figure*}[h!]
\vskip -0.1in
    \centering
    \includegraphics[width=0.9\textwidth]{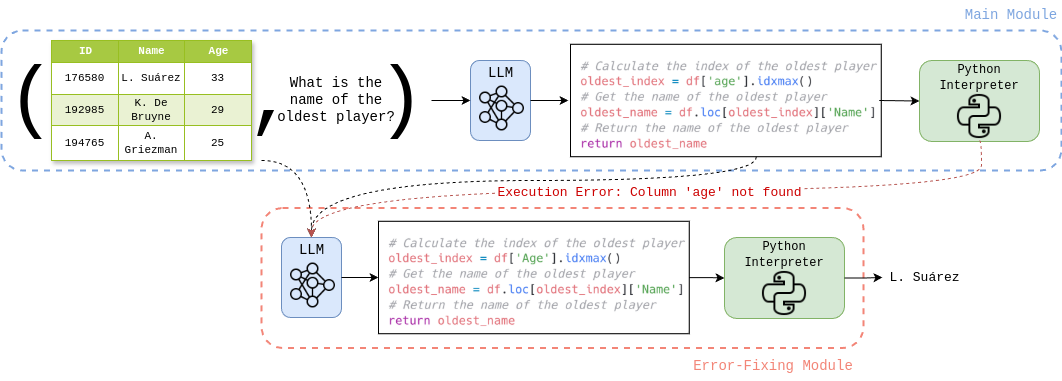}
    \caption{Overview of the architecture of our system.}
    \label{fig:vol2}
\end{figure*}

\section{System Overview}
Our system mainly performs text-to-Python code conversions to queries via prompting. 
This allows for a lightweight and task-agnostic implementation that employs LLMs without further training. Specifically, the system consists of two modules: the Main Module and the Error-Fixing Module. An overview of the architecture provided in Figure \ref{fig:vol2}.

\subsection{Main Module}
This module is responsible for translating the query $(T, Q)$ into executable Python code. First, a prompt is generated by applying a predefined template to the input  $(T, Q)$. This prompt is then provided to an LLM, which completes a Python function $P$ that takes the table $T$ as a \texttt{DataFrame} input and, when executed, answers the question $Q$. The generated code is executed by a Python interpreter, which calls the function with the table $T$ and outputs the predicted answer $\hat{a}$. If execution completes without errors, $\hat{a}$ is considered the system's final prediction. Otherwise, the Error-Fixing Module is invoked.

\paragraph{Prompt} 
The prompt template $f$ is used to structure the input of the LLM based on a $(T, Q)$ pair, incorporating the following fields:

\textbf{1. Task Description (TD):} A comment describing the objective of completing a Python function to answer $Q$, followed by the function header.

\textbf{2. Column Description (CD):} A multiline comment listing the column indices, names, data types, and example values in CSV format.

\textbf{3. Sample Rows (SR):} A comment containing the first $n$ rows of the table. Columns with values exceeding $n_c$ characters are truncated, and values are right-aligned with spaces \cite{rajkumar2022evaluatingtexttosqlcapabilitieslarge}.

\textbf{4. Columns Used and Answer Type (CUAT):} Comments specifying the columns required to answer $Q$, their data types, and the expected answer type.

Since CUAT is unknown for a new $(T, Q)$ pair, generating $P$ involves two steps. First, the LLM is prompted with $(T, Q)$ using an incomplete version of $f$, denoted as $f_{\text{inc}}$, which contains TD, CD, and SR, to infer CUAT. Then, the inferred CUAT is incorporated into $f$, forming a complete prompt with all four fields.

In both steps, we leverage in-context learning by incorporating $k$ manually crafted examples $\{(T_i, Q_i, P_i)\}_{i=1}^k$, where each $(T_i, Q_i)$ pair represents a table and its corresponding query, and $P_i$ is a program that processes $T_i$ to answer $Q_i$. Each example is structured using the same template $f$, (CUAT is extracted by the $P_i$), and is appended with the corresponding $P_i$. The complete prompt for the first step is:

$p_1 = (f(T_1, Q_1) || P_1) || \ldots || (f(T_k, Q_k) || P_k)\allowbreak || f_{\text{inc}}(T, Q)$
while for the second:
$p_2 = (f(T_1, Q_1) || P_1) || \ldots || (f(T_k, Q_k) || P_k) \allowbreak|| f(T, Q)$, where $p_1$, $p_2$ are the prompts used for Steps 1 and 2, respectively, and the CUAT field in $f(T, Q)$ of $p_2$ is computed by $g(p_1)$, where $g$ is the output of LLM given a prompt. 
Further details on prompt construction are provided in the Appendix \ref{app:prompts}. 

 

\paragraph{Example Programs}
Furthermore, to enhance program generation, we incorporate a CoT strategy. This involves inserting comments describing intermediate reasoning steps, interleaved with the Python statements implementing these steps. Our approach aligns with methods used in PAL \cite{gao2023palprogramaidedlanguagemodels} and PoT \cite{chen2022program}, aiming to leverage LLMs' reasoning capabilities while mitigating computational inaccuracies by offloading calculations to the Python interpreter. Meaningful variable names are also included in example programs, as they can further facilitate reasoning by clarifying intent
.

Upon observing the example programs, the LLM generates a completion of the Python function $P$ to answer $Q$, incorporating its reasoning in comments. The generated program is then executed by the Python interpreter, which outputs the predicted answer $\hat{a}$. If execution timeouts or fails, both the generated code and a description of the error are passed to the Error-Fixing Module.

\subsection{Error-Fixing Module}
This module utilizes an LLM prompted to correct execution errors in the function $P$. The prompt includes the following components:

\textbf{1. Task Description:} A description of the objective, instructing the model to fix the error.

\textbf{2. Faulty Program:} The erroneous code presented in the format $f(T, Q) || P$. 

\textbf{3. Code Error:} A description message of the execution error.

\textbf{4. Function Header:} The corrected function signature that the model will complete.

Once the LLM generates the corrected function, it is executed in the Python interpreter to  produce the answer $\hat{a}$. If execution completes successfully without timing out or failing, the system's final prediction is $\hat{a}$. Otherwise, it retries until the maximum attempts and outputs 
\texttt{Error}
if all fail.

\section{Experimental Setup}

\paragraph{Dataset}  
The dataset used for model development is publicly available on Hugging Face\footnote{\url{https://huggingface.co/datasets/cardiffnlp/databench}}. It contains 1,308 questions across 65 tables. We derive our exemplars from the predefined \texttt{train} split, comprising 988 questions from 49 tables. The columns required to answer each question, their data types, and the answer's data type  are precomputed by the task organizers and included in the dataset. Our annotations focus solely on the expected code and reasoning steps generated by the LLM. Model evaluation is conducted on the \texttt{dev} split, which includes 320 questions from 16 tables, as well as on the \texttt{test} set, which consists of 522 questions from 15 tables and serves as the basis for the final ranking.

\paragraph{Evaluation Metric}  
Model performance is assessed using a relaxed accuracy measure, which accounts for predictions with equivalent semantic meaning, e.g. the prediction \texttt{"Yes"} is considered correct if the gold answer is \texttt{True}. Numerical answers are truncated to two decimal places before being compared against the gold answer. The evaluation metric is provided by the task organizers through the \texttt{databench\_eval}\footnote{\url{https://github.com/jorses/databench_eval}} Python package.

\paragraph{Baseline}  
The baseline system employs a model to convert the input query into a Python function that takes the table as a \texttt{Pandas DataFrame} and returns the answer to the question. The generated function is then executed by the Python interpreter. The prompt includes detailed task instructions along with a simple example of a completed function. Additionally, it provides the names of the columns in the input \texttt{DataFrame} as a list. 
The model utilized is stable-code-3b-GGUF\footnote{\url{https://huggingface.co/TheBloke/stable-code-3b-GGUF}}.

\paragraph{Models and Hyperparameters}  
We evaluate the performance of multiple LLMs as components of our system. In all experiments, the same LLM is used for both the Main and Error-Fixing Modules. The models include open-source variants of Llama (8B, 70B, and 405B), proprietary models like Claude 3.5 Sonnet, and code-generation models such as Qwen 2.5-Coder 7B. Details on model versions are in Appendix \ref{app:models}.

For code generation, we select a temperature of \texttt{temp=0} and \texttt{top\_p=0.9} in all experiments, except for the Error-Fixing Module for smaller models ($\leq 70$B parameters), where we set  \texttt{temp=1} and allow up to three invocation attempts. The Main Module processes five rows from the table, while the Error-Fixing Module processes ten rows for both DataBench and DataBench Lite. The prompt includes nine exemplar cases, ensuring coverage of all possible answer data types in DataBench.

\paragraph{Resources}  
Experiments involving Claude 3.5 Sonnet and Llama 3.1 Instruct 405B are conducted using Amazon Bedrock, while experiments with the remaining models are performed using \href{https://ollama.com/}{Ollama} on 4 NVIDIA A10G GPUs.

\section{Results}  
\paragraph{Rankings}  
Table \ref{tab:main_results} presents accuracy scores on \texttt{dev} and \texttt{test} set for our system, considering LLMs of different sizes.
We rank $1^{st}$ in both subtasks in the proprietary models track when employing Claude 3.5 Sonnet, 
which demonstrates substantial improvements over the baseline, surpassing it by $\sim$60\% in both subtasks. 
Moreover, our system ranks $3^{rd}$ in DataBench in the open-source category using Llama 3.1 Instruct 405B.
Our system significantly outperforms the baseline even when employing smaller models ($\leq 8$B parameters). The accuracy scores of both smaller models rank just below the $2^{nd}$ and $3^{rd}$ positions in DataBench and DataBench Lite respectively, in the smaller model ranking. However, due to submission limitations in the official competition,  smaller models' predictions were not submitted and are therefore not reflected in the final competition rankings.

\begin{table*}[h!]
    \centering \small
    \begin{tabular}{c|c|c|c|c}
    \hline
    \multirow{2}{*}{Models} & \multicolumn{2}{c}{DataBench} & \multicolumn{2}{c}{DataBench Lite} \\ \cline{2-5}
    & Dev Set Accuracy & Test Set Accuracy & Dev Set Accuracy & Test Set Accuracy  \\
    \hline
    Claude 3.5 Sonnet & 91.87 & \textbf{85.63 (89.85)}  & 87.81 & \textbf{87.93 (88.89)}\\
    Llama 3.1 Instruct 405B & 91.56 & \textbf{83.33 (87.16)} & 90.00 & \textbf{77.78 (78.54)}\\
    \hline
    Llama 3.3 70B & 89.06 & 79.50 & 86.88 & 83.33\\
    \hline
    Llama 3.1 8B & 79.06 & 65.13 & 74.38 & 67.82 \\
    Qwen2.5-Coder 7B & 77.5 & 65.33 & 79.69 & 68.20 \\
    \hline
    Baseline & - & 26.00 & - & 27.00 \\
    \hline
    \end{tabular}
    \caption{Accuracy scores (\%) of our system across different LLMs on both subtasks. \textbf{Bold} denotes the submitted predictions and in the parenthesis are the official accuracy scores of those on DataBench after human reviewing.}
    \label{tab:main_results}
\end{table*}
\begin{table*}[h!]
    \centering \small
    \begin{tabular}{c|c|c|c|c}
    \hline
    \multirow{2}{*}{Prompt} & \multicolumn{2}{c}{DataBench} & \multicolumn{2}{c}{DataBench Lite} \\ \cline{2-5}
    & Dev Set Accuracy & Test Set Accuracy & Dev Set Accuracy & Test Set Accuracy  \\ \hline
    TD & 48.44 & 47.70 & 48.44 & 48.08\\
    TD + CD & 46.25 & 47.89 & 49.06 & 44.69 \\
    TD + CD + SR & 40.00 & 48.08 & 41.25 & 48.28 \\
    Simple Example + TD + CD + SR & 53.45 & 58.43 & 57.5 & 60.92 \\
    Few Shot & 71.88 & 64.18 & 73.12 & 65.33 \\
    Few Shot CoT & 75.31 & 65.71 & 79.69 & 67.43 \\
    Few Shot TD + CD + SR & 85.94 & 78.93 & 76.05 & 74.52 \\
    Few Shot CoT + TD + CD + SR  & 91.25 & 79.50 & 80.31 & 76.05 \\
    Main Module  & 89.38 & 82.57 &79.06 & 71.65 \\
    \hline
    
    \end{tabular}
    \caption{Accuracy scores (\%) for Llama 3.1 Instruct 405B with ablated prompts.}
    \label{tab:ablation}
\end{table*}
\begin{table*}[h!]
\centering \small
    \begin{tabular}{c|c|c|c|c|c|c}
    \hline
    \multirow{2}{*}{Model} & \multicolumn{3}{c}{DataBench - Test Set Accuracy} & \multicolumn{3}{c}{ DataBench Lite - Test Set  Accuracy}  \\ \cline{2-7}
    & Main-Module & Full System & \#Errors Fixed & Main-Module & Full System & \#Errors Fixed \\ \hline
    Claude 3.5 Sonnet & \textbf{83.52} & \textbf{85.63} & 11/16 & \textbf{85.82} & \textbf{87.93} & 11/16 \\
    Llama 3.1 Instruct 405B & 82.57 & 83.33 & 4/10 & 71.65 & 77.78 & 32/55 \\
    Llama 3.3 70B & 77.97 & 79.5 & 8/22 & 81.61 & 83.33 & 9/20 \\
    Llama 3.1 8B & 62.84 & 65.13  & 12/42 & 64.94 & 67.82 & 15/50 \\
    Qwen2.5-Coder 7B & 63.22 & 65.33 & 11/49 & 65.90 & 68.20 & 12/44\\
      
    \hline
    
    \end{tabular}
    \caption{Accuracy scores (\%) for the \texttt{test} set of Main Module and Full System, and number of errors fixed by the Error-Fixing Module for all models. \textbf{Bold} highlights best accuracy results across models.}
    \label{tab:error_fixing_module_significance}
\end{table*}
\paragraph{Large vs. Small Models}  
Our findings indicate that our system's  performance is heavily influenced by the overall capabilities of the LLM used. Larger models achieve up to 20\% higher accuracy in both subtasks, which is expected given their superior code generation capabilities. This is evident even when compared with a smaller model that is specifically trained for code generation tasks.

\paragraph{Information in the Prompt}  
We conduct an \textit{ablation study} on the different fields included in the prompt used in the Main Module to assess their impact on performance. Specifically, we evaluate the accuracy of the Main Module using Llama 3.1 Instruct 405B as the underlying LLM, incrementally increasing the amount of information provided in the prompt until it matches the final system configuration. The results are reported in Table \ref{tab:ablation}, including the following configurations:

\textbf{1. Simple Example + TD + CD + SR}: 
The prompt includes the fields TD, CD, and SR, along with a simple example of a completed function that calculates the number of rows in a \texttt{DataFrame}, as used in the baseline.

\textbf{2. Few-Shot (FS)}: The prompt includes examples of completed functions using single-line \texttt{Pandas} statements. The LLM is instructed to generate function completions in a similar format.

\textbf{3. FS CoT}: The prompt includes examples of completed functions with the reasoning steps in comments. The LLM is instructed to generate responses following the same approach.

\textbf{4. FS TD + CD + SR}: The prompt includes TD, CD, and SR, along with  examples of completed functions using single-line \texttt{Pandas} statements.
The LLM is instructed to generate single-line function completions.

\textbf{5. FS CoT + TD + CD + SR}: The prompt includes all fields of the Main Module, excluding CUAT. The exemplars contain reasoning steps as comments, and the LLM is prompted to generate responses following the same approach.

We observe that the sole inclusion of column and row descriptions provides minimal improvement in model accuracy on the \texttt{test} set, while performance decreases on the \texttt{dev} set. However, their significance becomes apparent when multiple exemplars illustrating the task are included in the prompt. In the FS setting, model performance significantly improves across all datasets and subtasks; 
Notably, the mere incorporation of a single task demonstration increases accuracy by more than 10\% across all datasets and subtasks.
Finally, allowing the LLM to explicitly express its reasoning—detailing the information extracted from the table, the expected outcome, and the intermediate steps to derive the answer—leads to a consistent improvement in performance across all tasks.

\paragraph{Significance of the Error-Fixing Module}
We present the \texttt{test} set accuracy scores with and without the involvement of the Error-Fixing Module, along with the number of correctly fixed errors—resulting in the gold label— in Table \ref{tab:error_fixing_module_significance}. The Error-Fixing Module provides an average accuracy increase of $2.4\%$ across all models and subtasks.  

Larger models, such as Claude 3.5 Sonnet and Llama 3.1 Instruct 405B, produce fewer execution errors and successfully correct more than $40\%$ of them across all tasks. In contrast, smaller models generate a higher number of errors and demonstrate limited error-correction capabilities, successfully resolving only $30\%$ of their execution errors at most. While further improvements to the Error-Fixing Module could substantially enhance the accuracy of smaller models, their impact on larger models would likely be minimal, as these models are inherently less prone to generating code that results in execution errors.

\section{Conclusion}
In this work, we present our system for SemEval-2025 Task 8: Question-Answering Over Tabular Data, which translates natural language questions into executable Python functions using only LLM prompting. Our approach decomposes the task into smaller steps, explicitly demonstrating its reasoning process, and incorporates a self-correction mechanism to handle execution errors. As a result, our system ranks first in both subtasks within the proprietary model category. 

\bibliography{main}

\begin{thebibliography}{22}
\providecommand{\natexlab}[1]{#1}

\bibitem[{Agarwal et~al.(2025)Agarwal, S, and Devaguptapu}]{agarwal2025hybridgraphstableandtextbased}
Ankush Agarwal, Ganesh S, and Chaitanya Devaguptapu. 2025.
\newblock \href {https://arxiv.org/abs/2501.17767} {Hybrid graphs for table-and-text based question answering using llms}.
\newblock \emph{Preprint}, arXiv:2501.17767.

\bibitem[{Chen(2023)}]{chen-2023-large}
Wenhu Chen. 2023.
\newblock \href {https://doi.org/10.18653/v1/2023.findings-eacl.83} {Large language models are few(1)-shot table reasoners}.
\newblock In \emph{Findings of the Association for Computational Linguistics: EACL 2023}, pages 1120--1130, Dubrovnik, Croatia. Association for Computational Linguistics.

\bibitem[{Chen et~al.(2022)Chen, Ma, Wang, and Cohen}]{chen2022program}
Wenhu Chen, Xueguang Ma, Xinyi Wang, and William~W Cohen. 2022.
\newblock Program of thoughts prompting: Disentangling computation from reasoning for numerical reasoning tasks.
\newblock \emph{arXiv preprint arXiv:2211.12588}.

\bibitem[{Cheng et~al.(2023)Cheng, Xie, Shi, Li, Nadkarni, Hu, Xiong, Radev, Ostendorf, Zettlemoyer, Smith, and Yu}]{cheng2023bindinglanguagemodelssymbolic}
Zhoujun Cheng, Tianbao Xie, Peng Shi, Chengzu Li, Rahul Nadkarni, Yushi Hu, Caiming Xiong, Dragomir Radev, Mari Ostendorf, Luke Zettlemoyer, Noah~A. Smith, and Tao Yu. 2023.
\newblock \href {https://arxiv.org/abs/2210.02875} {Binding language models in symbolic languages}.
\newblock \emph{Preprint}, arXiv:2210.02875.

\bibitem[{Fang et~al.(2024)Fang, Xu, Tan, Zhang, Hu, Qi, Nickleach, Socolinsky, Sengamedu, and Faloutsos}]{fang2024largelanguagemodelsllmstabular}
Xi~Fang, Weijie Xu, Fiona~Anting Tan, Jiani Zhang, Ziqing Hu, Yanjun Qi, Scott Nickleach, Diego Socolinsky, Srinivasan Sengamedu, and Christos Faloutsos. 2024.
\newblock \href {https://arxiv.org/abs/2402.17944} {Large language models(llms) on tabular data: Prediction, generation, and understanding -- a survey}.
\newblock \emph{Preprint}, arXiv:2402.17944.

\bibitem[{Gao et~al.(2023)Gao, Madaan, Zhou, Alon, Liu, Yang, Callan, and Neubig}]{gao2023palprogramaidedlanguagemodels}
Luyu Gao, Aman Madaan, Shuyan Zhou, Uri Alon, Pengfei Liu, Yiming Yang, Jamie Callan, and Graham Neubig. 2023.
\newblock \href {https://arxiv.org/abs/2211.10435} {Pal: Program-aided language models}.
\newblock \emph{Preprint}, arXiv:2211.10435.

\bibitem[{Gemmell and Dalton(2023)}]{gemmell2023generatetransformanswerquestion}
Carlos Gemmell and Jeffrey Dalton. 2023.
\newblock \href {https://arxiv.org/abs/2303.10138} {Generate, transform, answer: Question specific tool synthesis for tabular data}.
\newblock \emph{Preprint}, arXiv:2303.10138.

\bibitem[{Giang et~al.(2024)Giang, Nguyen, Brugere, Sharma, Kariyappa, Nguyen, and Lecue}]{giang2024interpretablellmbasedtablequestion}
Giang, Nguyen, Ivan Brugere, Shubham Sharma, Sanjay Kariyappa, Anh~Totti Nguyen, and Freddy Lecue. 2024.
\newblock \href {https://arxiv.org/abs/2412.12386} {Interpretable llm-based table question answering}.
\newblock \emph{Preprint}, arXiv:2412.12386.

\bibitem[{Hegselmann et~al.(2022)Hegselmann, Buendia, Lang, Agrawal, Jiang, and Sontag}]{Hegselmann2022TabLLMFC}
Stefan Hegselmann, Alejandro Buendia, Hunter Lang, Monica Agrawal, Xiaoyi Jiang, and David~A. Sontag. 2022.
\newblock \href {https://api.semanticscholar.org/CorpusID:252992811} {Tabllm: Few-shot classification of tabular data with large language models}.
\newblock \emph{ArXiv}, abs/2210.10723.

\bibitem[{Lei et~al.(2023)Lei, Luo, Yang, Liu, Liu, Lei, Huang, Wei, He, Zhao, and Liu}]{lei2023tableqakitcomprehensivepracticaltoolkit}
Fangyu Lei, Tongxu Luo, Pengqi Yang, Weihao Liu, Hanwen Liu, Jiahe Lei, Yiming Huang, Yifan Wei, Shizhu He, Jun Zhao, and Kang Liu. 2023.
\newblock \href {https://arxiv.org/abs/2310.15075} {Tableqakit: A comprehensive and practical toolkit for table-based question answering}.
\newblock \emph{Preprint}, arXiv:2310.15075.

\bibitem[{Liu et~al.(2022)Liu, Chen, Guo, Ziyadi, Lin, Chen, and Lou}]{liu2022tapextablepretraininglearning}
Qian Liu, Bei Chen, Jiaqi Guo, Morteza Ziyadi, Zeqi Lin, Weizhu Chen, and Jian-Guang Lou. 2022.
\newblock \href {https://arxiv.org/abs/2107.07653} {Tapex: Table pre-training via learning a neural sql executor}.
\newblock \emph{Preprint}, arXiv:2107.07653.

\bibitem[{Nam et~al.(2024)Nam, Song, Park, Tack, Yun, Kim, Oh, and Shin}]{Nam2024TabularTL}
Jaehyun Nam, Woomin Song, Seong~Hyeon Park, Jihoon Tack, Sukmin Yun, Jaehyung Kim, Kyu~Hwan Oh, and Jinwoo Shin. 2024.
\newblock \href {https://api.semanticscholar.org/CorpusID:271915812} {Tabular transfer learning via prompting llms}.
\newblock \emph{ArXiv}, abs/2408.11063.

\bibitem[{Os{\'e}s~Grijalba et~al.(2024)Os{\'e}s~Grijalba, Ure{\~n}a-L{\'o}pez, Mart{\'i}nez~C{\'a}mara, and Camacho-Collados}]{oses-grijalba-etal-2024-question}
Jorge Os{\'e}s~Grijalba, L.~Alfonso Ure{\~n}a-L{\'o}pez, Eugenio Mart{\'i}nez~C{\'a}mara, and Jose Camacho-Collados. 2024.
\newblock \href {https://aclanthology.org/2024.lrec-main.1179/} {Question answering over tabular data with {D}ata{B}ench: A large-scale empirical evaluation of {LLM}s}.
\newblock In \emph{Proceedings of the 2024 Joint International Conference on Computational Linguistics, Language Resources and Evaluation (LREC-COLING 2024)}, pages 13471--13488, Torino, Italia. ELRA and ICCL.

\bibitem[{Rajkumar et~al.(2022)Rajkumar, Li, and Bahdanau}]{rajkumar2022evaluatingtexttosqlcapabilitieslarge}
Nitarshan Rajkumar, Raymond Li, and Dzmitry Bahdanau. 2022.
\newblock \href {https://arxiv.org/abs/2204.00498} {Evaluating the text-to-sql capabilities of large language models}.
\newblock \emph{Preprint}, arXiv:2204.00498.

\bibitem[{Wang et~al.(2025)Wang, Zhou, Song, Huang, Chen, Ma, and Zhang}]{wang2025understandingcharacteristicscodegeneration}
Zhijie Wang, Zijie Zhou, Da~Song, Yuheng Huang, Shengmai Chen, Lei Ma, and Tianyi Zhang. 2025.
\newblock \href {https://arxiv.org/abs/2406.08731} {Towards understanding the characteristics of code generation errors made by large language models}.
\newblock \emph{Preprint}, arXiv:2406.08731.

\bibitem[{Wei et~al.(2023)Wei, Wang, Schuurmans, Bosma, Ichter, Xia, Chi, Le, and Zhou}]{wei2023chainofthoughtpromptingelicitsreasoning}
Jason Wei, Xuezhi Wang, Dale Schuurmans, Maarten Bosma, Brian Ichter, Fei Xia, Ed~Chi, Quoc Le, and Denny Zhou. 2023.
\newblock \href {https://arxiv.org/abs/2201.11903} {Chain-of-thought prompting elicits reasoning in large language models}.
\newblock \emph{Preprint}, arXiv:2201.11903.

\bibitem[{Wu and Hou(2025)}]{wu-hou-2025-efficient}
Jie Wu and Mengshu Hou. 2025.
\newblock \href {https://aclanthology.org/2025.coling-main.663/} {An efficient retrieval-based method for tabular prediction with {LLM}}.
\newblock In \emph{Proceedings of the 31st International Conference on Computational Linguistics}, pages 9917--9925, Abu Dhabi, UAE. Association for Computational Linguistics.

\bibitem[{Ye et~al.(2023)Ye, Hui, Yang, Li, Huang, and Li}]{10.1145/3539618.3591708}
Yunhu Ye, Binyuan Hui, Min Yang, Binhua Li, Fei Huang, and Yongbin Li. 2023.
\newblock \href {https://doi.org/10.1145/3539618.3591708} {Large language models are versatile decomposers: Decomposing evidence and questions for table-based reasoning}.
\newblock In \emph{Proceedings of the 46th International ACM SIGIR Conference on Research and Development in Information Retrieval}, SIGIR '23, page 174–184, New York, NY, USA. Association for Computing Machinery.

\bibitem[{Zha et~al.(2023)Zha, Zhou, Li, Wang, Huang, Yang, Yuan, Su, Li, Su, Zhang, Zhou, Shou, Wang, Zhu, Lu, Ye, Ye, Ye, Zhang, Deng, Xu, Wang, Chen, and Zhao}]{zha2023tablegptunifyingtablesnature}
Liangyu Zha, Junlin Zhou, Liyao Li, Rui Wang, Qingyi Huang, Saisai Yang, Jing Yuan, Changbao Su, Xiang Li, Aofeng Su, Tao Zhang, Chen Zhou, Kaizhe Shou, Miao Wang, Wufang Zhu, Guoshan Lu, Chao Ye, Yali Ye, Wentao Ye, Yiming Zhang, Xinglong Deng, Jie Xu, Haobo Wang, Gang Chen, and Junbo Zhao. 2023.
\newblock \href {https://arxiv.org/abs/2307.08674} {Tablegpt: Towards unifying tables, nature language and commands into one gpt}.
\newblock \emph{Preprint}, arXiv:2307.08674.

\bibitem[{Zhang et~al.(2023{\natexlab{a}})Zhang, Chang, and Ji}]{zhang2023bridginggapdecipheringtabular}
Hengyuan Zhang, Peng Chang, and Zongcheng Ji. 2023{\natexlab{a}}.
\newblock \href {https://arxiv.org/abs/2308.11891} {Bridging the gap: Deciphering tabular data using large language model}.
\newblock \emph{Preprint}, arXiv:2308.11891.

\bibitem[{Zhang et~al.(2023{\natexlab{b}})Zhang, Henkel, Floratou, Cahoon, Deep, and Patel}]{zhang2023reactableenhancingreacttable}
Yunjia Zhang, Jordan Henkel, Avrilia Floratou, Joyce Cahoon, Shaleen Deep, and Jignesh~M. Patel. 2023{\natexlab{b}}.
\newblock \href {https://arxiv.org/abs/2310.00815} {Reactable: Enhancing react for table question answering}.
\newblock \emph{Preprint}, arXiv:2310.00815.

\bibitem[{Zhu et~al.(2024)Zhu, Liu, Feng, Wang, Li, and Chua}]{10.1145/3677052.3698685}
Fengbin Zhu, Ziyang Liu, Fuli Feng, Chao Wang, Moxin Li, and Tat~Seng Chua. 2024.
\newblock \href {https://doi.org/10.1145/3677052.3698685} {Tat-llm: A specialized language model for discrete reasoning over financial tabular and textual data}.
\newblock In \emph{Proceedings of the 5th ACM International Conference on AI in Finance}, ICAIF '24, page 310–318, New York, NY, USA. Association for Computing Machinery.

\end{thebibliography}

\appendix

\section{Prompts}
\label{app:prompts}

\subsection{Main Module}  
We display the prompt used in the system's Main Module. A single annotated exemplar, i.e., $f(T_1,Q_1)||P_1$, is presented in Figure \ref{fig:prompt_annotated_exemplar}. Multiple exemplars are included before the definition of the function to be completed. The final segment of the prompt, $f_{\mathrm{inc}}(T, Q)$, corresponding to an example query, is shown in Figure~\ref{fig:prompt_query}.

\lstset{
    basicstyle=\small\ttfamily,  
    breaklines=true,             
}
\begin{figure*}
\begin{lstlisting}
# TODO: complete the following function. It should give the answer to: How many players have the position 'ST'?
def answer(df: pd.DataFrame):
  """
    #,Column,Non-Null CounT,Dtype,Types of Elements,Values
    0,ID,14620,uint32,[<class 'int'>],
    1,Name,14620,category,[<class 'str'>],5 example values are [' L. Suarez', ' K. De Bruyne', ' Bruno Fernandes', ' A. Griezmann', ' M. Acuna']
    2,Preferred Foot,14620,category,[<class 'str'>],All values are ['Right', 'Left']
    3,Position,14610,category,[<class 'str'>, nan],5 example values are ['RS', 'RCM', 'CAM', 'RW', 'LB'] 
    The first 5 rows from the dataframe:
           ID             Name  Preferred Foot  Position
    0  176580        L. Suarez           Right        RS
    1  192985     K. De Bruyne           Right       RCM
    2  212198  Bruno Fernandes           Right       CAM
    3  194765     A. Griezmann            Left        RW
    4  224334         M. Acuna            Left        LB
    """
  df.columns = ['ID', 'Name', 'Preferred Foot', 'Position']
  # The columns used to answer the question: ['Position']
  # The types of the columns used to answer the question: ['category']
  # The type of the answer: number
  # Create a boolean mask for rows where Position is 'ST'
  is_st_position = df['Position'] == 'ST'
  # Count the rows where the mask is True
  st_player_count = df[is_st_position].shape[0]
  # Return the count of players with 'ST' position
  return st_player_count
\end{lstlisting}    
\caption{Single annotated exemplar used in the prompt. Task description is directly sourced from the \texttt{databench\_eval} package provided by the organizers.}
\label{fig:prompt_annotated_exemplar}
\end{figure*}
\begin{figure*}
\begin{lstlisting}
# TODO: complete the following function. It should give the answer to: Is our average employee older than 35?
def answer(df: pd.DataFrame):
  """
    #,Column,Non-Null CounT,Dtype,Types of Elements,Values
    0,Age,1470,uint8,[<class 'int'>],
    1,Attrition,1470,category,[<class 'str'>],All values are ['Yes', 'No']
    2,BusinessTravel,1470,category,[<class 'str'>],All values are ['Travel_Rarely', 'Travel_Frequently', 'Non-Travel']
    3,DailyRate,1470,uint16,[<class 'int'>],
    4,Department,1470,category,[<class 'str'>],All values are ['Sales', 'Research & Development', 'Human Resources']
    The first 5 rows from the dataframe:
       Age  Attrition     BusinessTravel  DailyRate              Department
    0   41        Yes      Travel_Rarely       1102                   Sales
    1   49         No  Travel_Frequently        279  Research & Development
    2   37        Yes      Travel_Rarely       1373  Research & Development
    3   33         No  Travel_Frequently       1392  Research & Development
    4   27         No      Travel_Rarely        591  Research & Development
  """
  df.columns = ['Age', 'Attrition', 'BusinessTravel', 'DailyRate', 'Department']
  # The columns used to answer the question: 
\end{lstlisting}
\caption{The part of the prompt that includes the query $(T, Q)$, i.e. $f_{\mathrm{inc}}(T, Q)$}
\label{fig:prompt_query}
\end{figure*}

The column description notably includes (1) the number of non-null values present in each column, (2) the type of the column as parsed by \texttt{Pandas}, (3) a list of all Python types present in the column, and (4) example values of categorical columns enclosed in quotes to highlight cases requiring special handling. For instance, all categorical values in the column \texttt{'Name'} in Figure \ref{fig:prompt_annotated_exemplar} begin with a space character. In all our examples, we include up to five values. If all values in a categorical column are fewer than five, we explicitly note their values.  

The LLM is instructed, within the same prompt, to first predict the columns used to answer the question, their types, and the type of the answer. It then generates reasoning comments along with the code to derive the answer. The response of Llama 3.1 Instruct 405B for the prompt ending in Figure \ref{fig:prompt_query} is presented in Figure \ref{fig:main_module_response}. The model correctly identifies the relevant columns and the answer type. Subsequently, it generates the code in multiple steps, clearly outlining its reasoning through comments above the corresponding statements.

\begin{figure*}
\begin{lstlisting}
['Age']
  # The types of the columns used to answer the question: ['number[uint8]']
  # The type of the answer: boolean
  # Calculate the average age of the employees
  average_age = df['Age'].mean()  
  # Return True if the average age is greater than 35, False otherwise
  return average_age > 35
\end{lstlisting}
\caption{Example response from Llama 3.1 Instruct 405B as component of the Main Module.}
\label{fig:main_module_response}
\end{figure*}

\subsection{Error-Fixing Module}  
An example prompt for the Error-Fixing Module concerning an execution error is displayed in Figure \ref{fig:prompt_error_fixing}. The actual representation of roles in the implementation varies depending on the LLM used. The task description is followed by the function signature and the erroneous code generated by the LLM, $f(T, Q)||P$. The generation from the Main Module is then followed by the error message. The model is subsequently tasked with rewriting the function to both fix the execution error and correctly answer the question. The completion of Llama 3.1 Instruct 405B for the prompt in Figure \ref{fig:prompt_error_fixing} is shown in Figure \ref{fig:error_fixing_module_response}.  

The error message conveys critical information that the model could not infer from the initial prompt. For example, although the response of the Main Module in Figure \ref{fig:main_module_response} follows the essential steps to answer the question, the model is unaware that the column \texttt{'Weight Class'} contains the value \texttt{'Open'}. Given the potentially large size of tables in the dataset, crucial information necessary to answer the question may not be included in the prompt but can instead be inferred from the error message.

\begin{figure*}
    \begin{lstlisting}
-- SYSTEM --
You are an assistant tasked with helping a user fix a code error. The user has written a function that is supposed to answer a question about a table.
-- USER --
# Help me fix the code error of the following function by rewriting it. The function should return the answer to the question: Are there more than 100 lifters in the weight class someone that weights 82kg would compete in?
def answer(df: pd.DataFrame):
  """
    #,Column,Non-Null CounT,Dtype,Types of Elements,Values
    0,Lifter Name,...
    1,Age,...
    2,Weight Class,3000,category,[<class 'str'>],5 example values are ['59 kg', '83 kg', '105 kg', '66 kg', '74 kg']
    ...
  """
  df.columns = ...
  # The columns used to answer the question: ['Weight Class']
  # The types of the columns used to answer the question: ['category']
  # The type of the answer: boolean    
  # Find the weight class for someone who weighs 82kg
  weight_class = df['Weight Class'].unique()
  weight_class = [x for x in weight_class if int(x.split()[0]) >= 82]
  weight_class = min(weight_class)    
  # Count the number of lifters in that weight class
  num_lifters = df[df['Weight Class'] == weight_class].shape[0]    
  # Return True if there are more than 100 lifters, False otherwise
  return num_lifters > 100
# The function outputs the following error:
# __CODE_ERROR__: invalid literal for int() with base 10: 'Open'
-- ASSISTANT --
def answer(df: pd.DataFrame):
  """
    ...
  """
  df.columns = ...
    \end{lstlisting}
    \caption{The Error-Fixing Module Prompt for Llama Models. The special tokens used by the Llama Models when roles change are replaced by the name of the role between dashes.}
    
    \label{fig:prompt_error_fixing}
\end{figure*}

\begin{figure*}
\begin{lstlisting}
# Find the weight class for someone who weighs 82kg
  weight_class = df['Weight Class'].unique()
  weight_class = [x for x in weight_class if x.replace('kg', '').strip().isdigit() and int(x.replace('kg', '').strip()) >= 82]
  weight_class = min(weight_class, key=lambda x: int(x.replace('kg', '').strip()))
  # Count the number of lifters in that weight class
  num_lifters = df[df['Weight Class'] == weight_class].shape[0]  
  # Return True if there are more than 100 lifters, False otherwise
  return num_lifters > 100    
\end{lstlisting}
\caption{Example response from Llama 3.1 Instruct 405B as component of the Error-Fixing Module.}
\label{fig:error_fixing_module_response}
\end{figure*}

\subsection{Prompt Size}

\begin{table}[!ht]
    \centering\small
    \begin{tabular}{p{0.2\linewidth}|p{0.2\linewidth}|p{0.2\linewidth}|p{0.2\linewidth}}
    \hline 
        \multicolumn{2}{c}{M.M Avg. \#Tokens} & \multicolumn{2}{c}{E.F.M Avg. \#Tokens} \\ \hline
        DB & DB Lite & DB & DB Lite \\ \hline \hline
        \multicolumn{4}{c}{\textbf{Llama Family} (128K)}  \\ \hline \hline
        19,160.9 & 6,518.2 & 2.738.4 & 3268.2 \\ \hline \hline
        \multicolumn{4}{c}{\textbf{Claude 3.5 Sonnet} (200K)}  \\ \hline \hline
        22,764.7 & 7,391.7 & 6,573.4 & 4,037.8  \\ \hline \hline
        \multicolumn{4}{c}{\textbf{Qwen2.5-Coder 7B} (128K)} \\ \hline \hline
        21,816.1 & 7,058.2 & 2,700.2 & 3,117.7 \\ \hline
    \end{tabular}
    
    \caption{Average Token Count for the prompt generated in the Main Module (M.M) and Error-Fixing Module (E.F.M) for both subtasks of the \texttt{test} set for the different model families. The size of the context window for the different families is presented in parenthesis.}
    \label{tab:token_lengths}
\end{table}

We present the average number of tokens in the prompts for both the Main and Error-Fixing Modules across the different model families in Table \ref{tab:token_lengths}. On average, the Main Module utilizes approximately 20K for all models.  
Meanwhile, the Error-Fixing Module includes the header and the erroneous code produced in the prompt without additional exemplars demonstrating the task. As a result, the Error-Fixing Module requires fewer tokens, remaining below 4K tokens in most cases.

\section{Exploratory data analysis}
\label{sec:exploratory}

\paragraph{Tables}  

\begin{figure*}[h!]
\centering
\begin{subfigure}{.45\textwidth}
  \centering
  \includegraphics[width=\linewidth]{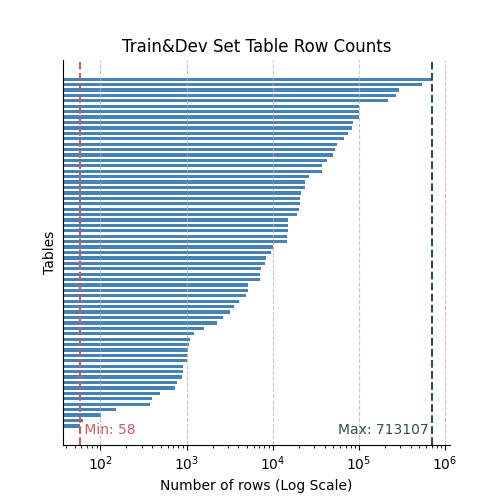}
\end{subfigure}%
\begin{subfigure}{.45\textwidth}
  \centering
  \includegraphics[width=\linewidth]{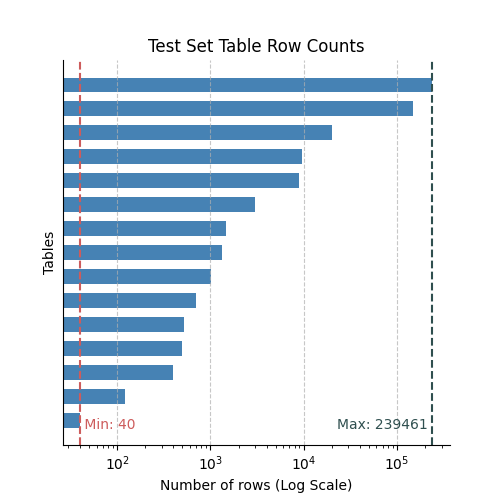}
\end{subfigure}
\caption{Number of rows of all tables in \texttt{train} \& \texttt{dev} set and in the \texttt{test} set.}
\label{fig:num_rows}
\end{figure*}

The DataBench dataset consists of a total of 80 tables (65 in the \texttt{train} and \texttt{dev} sets, and 15 in the \texttt{test} set) of varying sizes and diverse content. The distribution of table row counts across all sets in the DataBench task is shown in Figure \ref{fig:num_rows}. Table sizes range from fewer than 100 rows to several hundred thousand rows. In contrast, all tables in DataBench Lite contain only 20 rows, sampled from the larger dataset.  

\begin{figure*}[h!]
\centering
\begin{subfigure}{.45\textwidth}
  \centering
  \includegraphics[width=\linewidth]{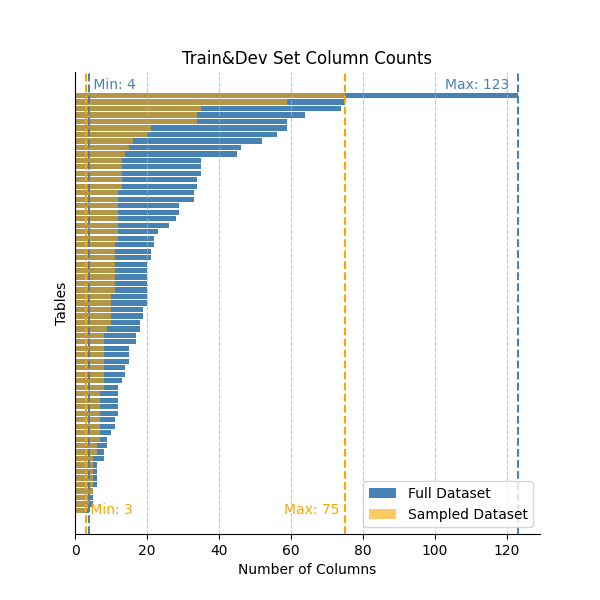}
\end{subfigure}%
\begin{subfigure}{.45\textwidth}
  \centering
  \includegraphics[width=\linewidth]{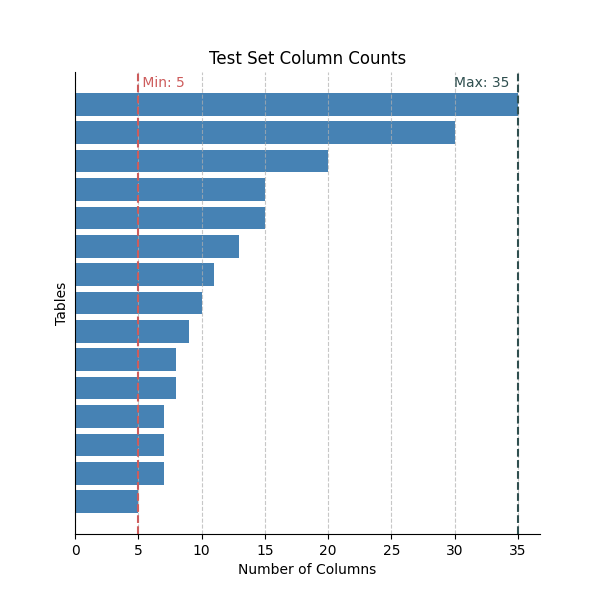}
\end{subfigure}
\caption{Number of columns of all tables in \texttt{train} \& \texttt{dev} set and in the \texttt{test} set. Sampled versions of the tables used in the \texttt{test} set for DataBench Lite include all columns of the original tables.}
\label{fig:num_columns}
\end{figure*}

Figure \ref{fig:num_columns} presents the distribution of column counts across all dataset splits. The number of columns in DataBench ranges from 3 to 123 in the \texttt{dev} set and up to 35 in the \texttt{test} set.

\paragraph{Questions}

DataBench and DataBench Lite contain the same set of questions, with the only difference being that DataBench Lite uses sampled subsets of the original tables. Example questions, along with their corresponding answers, answer data types, and the columns used in their derivation, are presented in Table \ref{tab:sample_dataset_rows}. All questions are answered exclusively using information present in the table.

\begin{figure}[t!]
    \centering
    \includegraphics[width=0.9\linewidth]{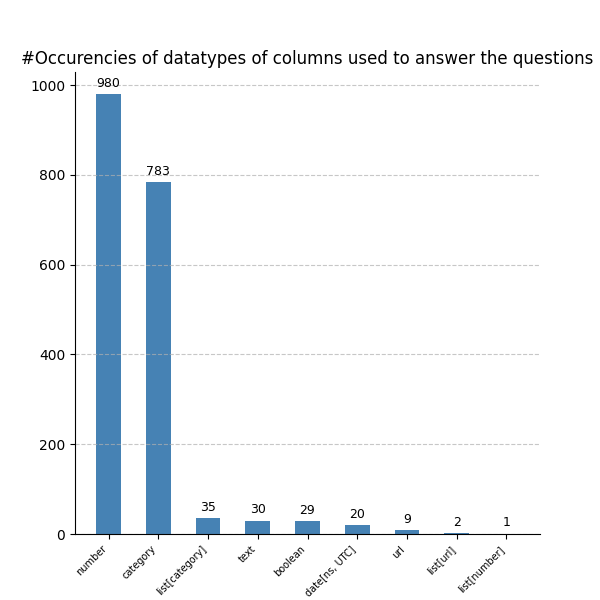}
    \caption{Number of times a column of each data type is used in a query. This categorization includes the samples in the \texttt{train} and \texttt{dev} set, which are annotated by the task organizers.}
    \label{fig:columns_used}
\end{figure}

In the annotated \texttt{train} and \texttt{dev} sets, the derivation of answers involves one, two, or three columns. Figure \ref{fig:columns_used} presents an overview of the column data types utilized across the dataset. The most commonly used column data types in these sets are \texttt{number} and \texttt{category}.

The dataset defines five possible answer data types:
\begin{enumerate} 
    \item \texttt{boolean}: The only type with a predefined set of values (\texttt{True} or \texttt{False}).
    \item \texttt{number}: Includes both integers and real numbers.
    \item \texttt{category}: A single categorical value represented as a string.
    \item \texttt{list[category]}: A list containing a fixed number of categorical values.
    \item \texttt{list[number]}: A list containing a fixed number of numerical values.
\end{enumerate}

Both datasets contain an equal number of questions corresponding to each possible answer data type.

\paragraph{Use of Textual Columns}

We analyzed the role of textual columns in cases where they contribute to the answer in the annotated \texttt{train} and \texttt{dev} sets, identifying 30 such instances. These cases include questions where the \texttt{text} data type appears among the utilized column data types included in the annotations. Textual columns are used in one of the following ways:

\begin{enumerate}
    \item \textbf{Uniqueness Tests}: The answer depends on whether the column contains unique values.
    \item \textbf{Length Conditions}: The length of the text in each cell contributes to the answer.  
    \item \textbf{Existence and Occurrence of Substrings}: The answer is derived based on whether a cell contains a specific substring or the number of times the substring appears within the cell.
    \item \textbf{Word Count}: The number of words in each cell contributes to the answer.
    \item \textbf{Index-Based Lookup}: The answer consists of one or more cells from the column, selected based on a Boolean mask.
\end{enumerate}

We observe that all of these tasks can be addressed solely through programmatic statements and do not require more advanced natural language processing capabilities, such as sentiment analysis of each cell of a textual column. Consequently, we focus exclusively on program generation and do not explore more sophisticated techniques for extracting information from textual columns, such as those proposed in \cite{cheng2023bindinglanguagemodelssymbolic, gemmell2023generatetransformanswerquestion}.


\begin{figure}[t!]
    \centering
    \includegraphics[width=0.9\linewidth]{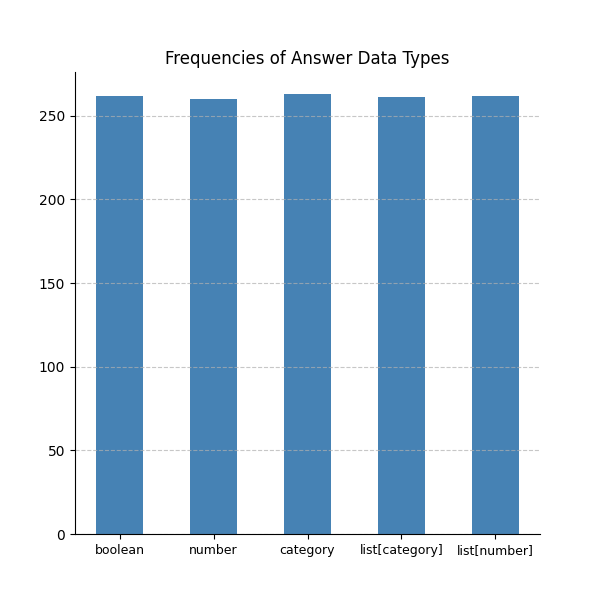}
    \caption{The frequencies of every answer data type in DataBench \texttt
    {train} and \texttt{dev} set.}
    \label{fig:type_answer}
\end{figure}
\begin{table*}[]
    \centering \small
    \begin{tabular}{p{0.3\textwidth}|p{0.15\textwidth}|c|p{0.25\textwidth}}
    \hline
         Question & Answer & Answer Type & Columns used \\ \hline \hline
         How many unique types of animals are there? & \texttt{4} & \texttt{number} & \texttt{['class\_type']} \\ \hline
         Is the maximum level of Extraversion greater than the maximum level of Agreeableness?	& \texttt{True} &	\texttt{boolean} &	\texttt{['Extraversion', 'Agreeableness']} \\ \hline
         What are the top 3 reviewer locations with the most reviews? &	\texttt{['United States', 'Australia', 'Malta']} &	\texttt{list[category]} & \texttt{['Reviewer\_Location']}
\\ \hline
        What is the most frequent age group among the respondents? & \texttt{25-34}	& \texttt{category} & \texttt{['How old are you?']}\\ \hline
        What are the top 4 numbers of claims in the patents? &	\texttt{[12, 18, 7, 13]} &	\texttt{list[number]} &	\texttt{['num\_claims']} \\ \hline

    \end{tabular}
    \caption{Sample questions with the corresponding answers, data types of answer and the names of the columns used to answer the question sampled from DataBench Lite \texttt{train} and \texttt{dev} sets.}
    \label{tab:sample_dataset_rows}
\end{table*}

\section{Example Generations of the Main Module}  

Responses generated by the Main Module for the questions in Table \ref{tab:sample_dataset_rows} are presented in Table \ref{tab:main_module_responses}. The responses were produced using Claude 3.5 Sonnet. The model first predicts the columns it will use to answer the question, along with the expected data type of the answer. It then generates the function completion, demonstrating a strong ability to follow exemplars and decompose its reasoning into smaller steps. This structured thought process allows for efficient verification and provides some confidence in assessing the correctness of the generated response.  

Larger models, such as Claude 3.5 Sonnet and Llama 3.1 Instruct 405B, exhibit a strong grasp of the underlying programming language, making them less prone to syntactic errors \cite{wang2025understandingcharacteristicscodegeneration}. However, they are still susceptible to semantic errors such as reasoning mistakes. For instance, in the first entry of Table \ref{tab:main_module_responses}, the model incorrectly predicts the relevant column to use, despite the column name being partially mentioned in the question. Additionally, LLMs can make more subtle reasoning errors. In the final entry, for example, the Main Module incorrectly retrieves the four highest values of the numerical column \texttt{num\_claims} instead of identifying the four most common values, as required.  

Even in cases where errors occur, the structured format of the generated responses allows for quick validation of the LLM's output. The categorization of errors in code generation has been an area of recent research \cite{wang2025understandingcharacteristicscodegeneration}. Adapting these error taxonomies to the tabular setting and further refining error classification is a promising research direction. Such efforts could enhance our understanding of errors LLMs make when translating natural language queries to code in Table Question Answering (TQA) and aid in developing methods to mitigate them.

\begin{table*}[]
    \centering \small
    \begin{tabular}{p{0.8\textwidth}|p{0.15\textwidth}} \hline
    Main Module Response & Result\\ \hline \hline
    \begin{lstlisting}
    ['animal_name']
    # The types of the columns used to answer the question: ['category']
    # The type of the answer: number
    # Count the number of unique animal names
    unique_animals = df['animal_name'].nunique()
    # Return the count of unique animals
    return unique_animals
    \end{lstlisting} & \textcolor{red}{Wrong Columns Used} \\ \hline
    \begin{lstlisting}
    ['Extraversion', 'Agreeableness']
    # The types of the columns used to answer the question: ['number[double]', 'number[double]']
    # The type of the answer: boolean
    # Get the maximum values for both Extraversion and Agreeableness
    max_extraversion = df['Extraversion'].max()
    max_agreeableness = df['Agreeableness'].max()
    # Compare the maximum values and return True if max_extraversion is greater than max_agreeableness
    return max_extraversion > max_agreeableness
    \end{lstlisting} & \textcolor{green}{Correct Prediction}\\ \hline

    \begin{lstlisting}
    ['Reviewer_Location']
    # The types of the columns used to answer the question: ['category']
    # The type of the answer: list[category]
    # Count the number of reviews per location and get the top 3
    top_3_locations = df['Reviewer_Location'].value_counts().nlargest(3).index.tolist()
    return top_3_locations
    \end{lstlisting} & \textcolor{green}{Correct Prediction}\\ \hline

    \begin{lstlisting}
    ['How old are you?']
    # The types of the columns used to answer the question: ['category']
    # The type of the answer: category
    # Find the most frequent age group by using value_counts()
    most_frequent_age = df['How old are you?'].value_counts().index[0]
    # Return the most frequent age group
    return most_frequent_age
    \end{lstlisting} & \textcolor{green}{Correct Prediction}\\ \hline
    
    \begin{lstlisting}
    ['num_claims']
    # The types of the columns used to answer the question: ['number[uint8]']
    # The type of the answer: list[number]
    # Get the top 4 values from the 'num_claims' column
    top_4_claims = df['num_claims'].nlargest(4)
    # Return the top 4 values as a list
    return top_4_claims.tolist()
    \end{lstlisting} & \textcolor{red}{Semantic Error} \\ \hline
    
    \end{tabular}
    \caption{Responses of the Main Module  when queried on the questions of Table \ref{tab:sample_dataset_rows}. Claude 3.5 Sonnet is used as the underlying LLM. The first three lines of every response correspond to the completion of the CUAT field and the remaining to the code answering the question. }
    \label{tab:main_module_responses}
\end{table*}

\section{Model Versions}
\label{app:models}

The models used in this study are the latest available versions at the time of experimentation. Specifically, they include a combination of proprietary and open-source models, along with models trained specifically for code generation. The proprietary models include Claude 3.5 Sonnet\footnote{\url{anthropic.claude-3-5-sonnet-20241022-v2:0}}, developed by Anthropic. Additionally, we incorporate open-source models such as Llama 3.3 70B\footnote{\url{https://ollama.com/library/llama3.3}}, Llama 3.1 8B\footnote{\url{https://ollama.com/library/llama3.1}}, and Llama 3.1 Instruct 405B\footnote{\url{meta.llama3-1-405b-instruct-v1:0}}, which offer varying parameter sizes to assess performance scaling effects. Lastly, we include Qwen 2.5-Coder 7B\footnote{\url{https://ollama.com/library/qwen2.5-coder}}, a model specifically optimized for code generation, to evaluate its effectiveness in program synthesis tasks.

\end{document}